\documentclass[letterpaper, 10 pt, conference]{ieeeconf}
\IEEEoverridecommandlockouts
\usepackage{graphics} 
\usepackage{epsfig} 
\usepackage{subfigure}
\usepackage[utf8]{inputenc}
\usepackage{mathptmx} 
\usepackage{times} 
\usepackage{amsmath} 
\usepackage{amssymb}  
\usepackage{multirow}
\usepackage{bm} 
\usepackage{multicol} 
\usepackage{arydshln}
\usepackage{diagbox}
\usepackage{float}
\usepackage[ruled,linesnumbered]{algorithm2e}
\usepackage{tabularx}
\usepackage{multirow}
\usepackage{cite}
\usepackage[normalem]{ulem}
\usepackage{url}
\usepackage{hyperref}
\usepackage{algorithmic}
\usepackage{color}
\usepackage[utf8]{inputenc}

\begin{document}

\title{Safe Online Gain Optimization for Variable Impedance Control}

\author{Changhao Wang*$^{1}$, Zhian Kuang*$^{1,2}$, Xiang Zhang*$^{1}$, and Masayoshi Tomizuka$^1$
\thanks{* denotes equal contribution.}
\thanks{$^1$Changhao Wang, Zhian Kuang, Xiang Zhang, and Masayoshi Tomizuka are with the Department of Mechanical Engineering, University of California, Berkeley, CA 94720, USA. 
{\tt\small \{changhaowang, zhiankuang, xiang\_zhang\_98, tomizuka\}@berkeley.edu}}
\thanks{$^2$Zhian Kuang is also with the Research Institute of Intelligent Control and Systems, Harbin Institute of Technology, 150001, Harbin, P.R. China.
}}

\maketitle

\begin{abstract}
Smooth behaviors are preferable for many contact-rich manipulation tasks. Impedance control arises as an effective way to regulate robot movements by mimicking a mass-spring-damping system. Consequently, the robot behavior can be determined by the impedance gains.
However, tuning the impedance gains for different tasks is tricky, especially for unstructured environments. Moreover, online adapting the optimal gains to meet the time-varying performance index is even more challenging.
In this paper, we present Safe Online Gain Optimization for Variable Impedance Control (Safe OnGO-VIC). By reformulating the dynamics of impedance control as a control-affine system, in which the impedance gains are the inputs, we provide a novel perspective to understand variable impedance control. Additionally, we innovatively formulate an optimization problem with online collected force information to obtain the optimal impedance gains in real-time.
Safety constraints are also embedded in the proposed framework to avoid unwanted collisions.
We experimentally validated the proposed algorithm on three manipulation tasks. Comparison results with a constant gain baseline and an adaptive control method prove that the proposed algorithm is effective and generalizable to different scenarios. Experiment videos are available at
\href{https://msc.berkeley.edu/research/safe-ongo-vic.html}{https://msc.berkeley.edu/research/safe-ongo-vic.html}.
\end{abstract}

\section{Introduction}
Impedance control is a powerful way to let robots safely interact with unknown environments. However, it is challenging and time-consuming to tune the impedance gains in order to achieve desired robot behaviors. Furthermore, the adjusted gains may be specific to one environment and can hardly generalize to others. Therefore, a method that is able to obtain the optimal gains for different scenarios is vital for practical use of impedance control.

To adapt the optimal gains for different scenarios, adaptive control has been studied to adjust the controller online~\cite{muratore2019self,johannsmeier2019framework}. The impedance gains will adapt following a predefined adaptive control law.
However, to get a satisfactory result, it is necessary to tune good initial impedance gains. Moreover, most adaptive laws introduce additional hyper-parameters to the problem and complicate the overall tuning process.

Machine learning is a powerful way to learn the optimal impedance controller. Well-known methods include learning from demonstrations (LfD)~\cite{peternel2015human,abu2018force,zhang2021learning} and reinforcement learning (RL)~\cite{li2017adaptive, martin2019variable,rey2018learning,zhang2021learning}. For LfD approaches such as~\cite{peternel2015human, abu2018force}, the robot collects data from expert demonstrations, and a variable impedance policy can be obtained by fitting the collected data. However, the learned policy is specific to the demonstrated task and may not be robustly transferred to different task settings. 
In the RL field, by encoding the robot's performance into a reward function and setting variable impedance as the action space, the robot can learn the variable impedance policy to achieve the desired performance in the simulation~\cite{li2017adaptive,martin2019variable,rey2018learning}. However, as a common problem of machine learning-based methods, it is time-consuming to collect data and train the policy, which limits their ability for real-time application. 

Optimization is also a powerful tool for gain tuning. Mehdi~\cite{mehdi2011impedance} pioneered to use Particle Swarm Optimization (SWO) to tune the impedance controller offline. Similarly, Lahr~\cite{lahr2017adjustable} proposed to use a multi-objective genetic algorithm to optimize the impedance gains and demonstrated the effectiveness on a real robot. Unfortunately, these methods usually require extensive computation power. The high computation time limits their use in the online phase. One possible reason for the low efficiency is that they do not have an explicit relation between the impedance gains and the robot behavior. Thus, they had to rely on black-box optimization solvers to deal with an overly complicated problem. To the author's knowledge, there are no online optimization methods in the literature for tuning the impedance controller.

To obtain the optimal impedance gains in real-time, we propose an online gain optimization framework (Safe OnGO-VIC).
In contrast with other methods that view impedance gain as a parameter,
we express the dynamics of the impedance control into a control-affine system and consider the impedance gain as a control input.
By doing these,
we provide a new perspective to understand gain tuning and innovatively formulate an optimization problem to optimize the impedance gains online. An objective function that is able to regulate smooth robot behaviors is designed. Compared with previous optimization and learning methods, our method does not require offline data collection and can be optimized online for different tasks and environments efficiently. Furthermore, benefiting from the novel structure, safety constraints can be embedded into the framework in order to avoid unwanted collisions by adjusting the impedance gains.

In summary, the contributions of this paper are as follows:
\begin{itemize}
    \item A new perspective to understand the relationship between impedance gains and the robot states.
    \item An efficient online gain optimization framework for variable impedance control.
    \item Collision avoidance for variable impedance control.
    \item Comparative experiments demonstrating the effectiveness of the proposed algorithm.
\end{itemize}

\section{Proposed Method}
\label{PROPOSED APPROACH}

\subsection{Preliminary: Cartesian Space Impedance Control}
The dynamics model of a $6$-DOF manipulator in Cartesian space is written as~\cite{song2017impedance}
\begin{equation}
    M_r(p) \ddot{p} + C_r(p,\dot{p})\dot{p} +G_r(p) = J^{-T}\tau +F
    \label{eq:cartesian-dyanmics}
\end{equation}
where $\ddot{p}$, $\dot{p}$, $p\in \mathbb{R}^{6}$ are the Cartesian acceleration, velocity and position of the robot end-effector, $ M_r(p) \in \mathbb{R}^{6\times 6}$ stands for the mass-inertia matrix, $C_r(p,\dot{p}) \in \mathbb{R}^{6\times 6}$ denotes the Coriolis matrix, $G_r(p) \in \mathbb{R}^{6}$ is the gravity vector, $J\in \mathbb{R}^{6\times 6}$ is the Jacobian matrix, $F \in \mathbb{R}^{6}$ represents the external force, and, $\tau \in \mathbb{R}^{6}$ stands for the torque input of the joints.

By utilizing the impedance control law~\cite{song2017impedance}
\begin{align}
    \tau &= J^{T}\left\{ -F+ M_r(p)\ddot{p}_d+ C_r(p,\dot{p})\dot{p} +G_r (p)  \nonumber \right.  \\&~~
    \left. -M_r M^{-1}[K_d(\dot{p}-\dot{p_d})+ K_p(p-p_d) -F] \right\}\label{impdeancecontrollaw}
\end{align}
the robot system acts as a mass-spring-damping system
\begin{align} \label{impedance dynamics}
    M \ddot{e}+K_d \dot{e}+K_p e=F
\end{align}
where $e=p-p_d$, $\dot{e}=\dot{p}-\dot{p}_d$ and $\ddot{e}=\ddot{p}-\ddot{p}_d$ denote tracking position, velocity and acceleration errors. $M$, $K_d$ and $K_p\in \mathbb{R}^{6\times 6}$ are the desired mass, damping and stiffness matrices. 
By selecting $M$, $K_d$, and $K_p$ matrices, we can change the characteristics of the robot. 

\subsection{Dynamics Formulation}
\label{dynamics_section}

For impedance control~(\ref{impedance dynamics}), a necessary condition to ensure stability is to guarantee the impedance gains are positive-definite~\cite{kronander2016stability}. Therefore, we can left-multiply the inverse of $M$ and obtain
\begin{equation}
    \ddot e + K_d^\prime \dot e + K_p^\prime e = F^\prime
\label{eqn::dynamics_derivation_1}
\end{equation}
where $K_d^\prime = M^{-1} K_d$ and $K_p^\prime = M^{-1} K_p$ are the transformed gain matrices and $F^\prime = M^{-1} F$.

We assume that the gain matrices $M$, $K_d$, and $K_p$ are diagonal matrices~\cite{lawrence1988impedance, seraji1997force}
\begin{equation}
\begin{aligned}
    K_d^\prime & = diag\{ K_{d1}/M_{1},~\dots,~K_{d6} /M_{6}\} = diag\{K_{d1}^\prime,\dots,K_{d6}^\prime\} \\
    K_p^\prime & = diag\{ K_{p1}/M_1,~\dots,~K_{p6}/M_6\} = diag\{K_{p1}^\prime,\dots,K_{p6}^\prime\} \\
    F^\prime & = [ F_{1}/M_1,\dots, F_{6}/M_6]^T \nonumber 
\end{aligned}
\end{equation}
where $\{M_{1},\dots, M_{6}\},\{K_{d,1},\dots, K_{d6}\}, \{K_{p1},\dots, K_{p6}\}$ are diagonal elements of $M$, $K_d$, and $K_p$ matrices, and $F = [F_{1},\dots, F_{6}]^T$ is the 6-dimensional external wrench given by the environment. 
We can further reformulate (\ref{eqn::dynamics_derivation_1}) into a control-affine form by collecting terms together:
\begin{equation}
    \dot{x} = \begin{bmatrix}
    \dot{e} \\
    \ddot{e}
\end{bmatrix} = 
f(x) + g(x)u =
\begin{bmatrix}
    \dot{e} \\
    0
\end{bmatrix} +
\begin{bmatrix}
    0 \\
    g_2(x)
\end{bmatrix} u
\label{eqn::control_affine_dynamics}
\end{equation}
where the input $u = {[K_{d1}^\prime,\dots, K_{d6}^\prime,K_{p1}^\prime,\dots,K_{p6}^\prime,M_1^{-1},\dots,M_6^{-1}}]^T$. $g_2(x)$ is given by:
\begin{equation} \nonumber 
g_2(x) = 
\begin{bmatrix}
    \begin{array}{ccc|ccc|ccc}
-\dot{e}_{1} &       &           &-e_{1}  &           &       & F_{1} &       & \\
                &\ddots &           &       & \ddots    &       &   & \ddots& \\
                &       &-\dot{e}_{6}&       &           & -e_{6} &   &       & F_{6}
    \end{array}
\end{bmatrix}
\end{equation}

Equation (\ref{eqn::control_affine_dynamics}) reveals the relation between impedance gains and the robot states. With this specific formulation, we can optimize the gain based on these dynamics in order to regulate future robot trajectories.

In traditional methods, impedance gain is considered as a parameter instead of a control input when tuning the controller. To the authors' knowledge, this is the first method to formulate impedance gain as input and construct a dynamics system for gain tuning. By doing these, we illustrate a new way to understand the tuning process and provide convenience for the following optimization operations.

\subsection{Objective Function}
Minimizing the integral of time-weighted absolute error (ITAE)~\cite{martins2005tuning} is a commonly used objective to tune PID parameters offline: 
\begin{equation}
\begin{aligned}
    ITAE &= \int_{0}^{+\infty} t|e(t)| dt \\
\end{aligned}
\label{eqn::ITAE_loss}
\end{equation}
where $e(t)$ is the robot position error at time $t$. 

For contact-rich manipulation tasks, such as grinding and polishing, we prefer smoother trajectories. 
Therefore, inspired by the idea of the ITAE, we propose a new cost function named the finite-time integral of time-weighted absolute velocity error~(FITAVE) in (\ref{eqn::FITAVE_loss}) to minimize the velocity error for a smooth trajectory.

\begin{equation}
\begin{aligned}
    FITAVE &= \int_{0}^{T} t|\dot e(t)| dt \\
         &= \int_{0}^{T} t \bigg|\int_{0}^{t} \ddot e(\tau) d \tau \bigg| dt \\
         &= \int_{0}^{T} t \bigg| \int_{0}^{t} g_2(x) u d\tau \bigg| dt \\
\end{aligned} 
\label{eqn::FITAVE_loss}
\end{equation}
where $x$ is the robot states, $g_2(x)$ is the robot dynamics in~(\ref{eqn::control_affine_dynamics}), and $u$ is the optimization variable for impedance gains. 


\begin{figure}[http]
  \centering
   \includegraphics[width=250pt]{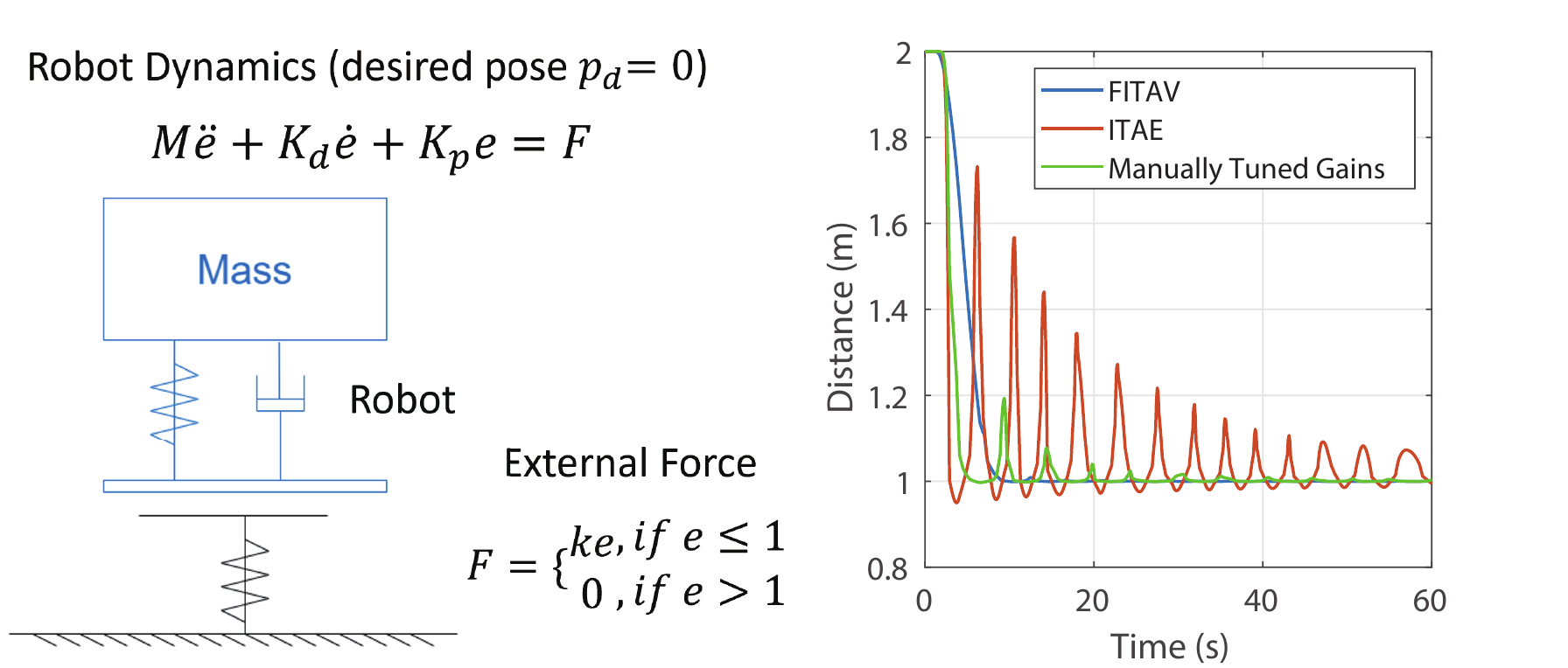}
   \caption{The simulation results of FITAVE}
   \label{fig:simu_FITAVE}
\end{figure}

Fig.~\ref{fig:simu_FITAVE} shows an example of FITAVE. A 1-DoF robot controlled by an impedance controller is going to contact a surface located at $e=1$. The dynamics of the surface is modeled by a spring with large stiffness. 
The objective is to select proper impedance gains to make contact as smooth as possible. We compared the results obtained from FITAVE with ITAE and manually tuned gains. The simulation result corroborates that FITAVE can generate smooth robot behaviors as we expected. Our experiment results in Section~\ref{exp2} further prove the effectiveness of FITAVE.

\subsection{Collision Constraints for Safe Interaction}
Consider a safe set defined as $h(x)\geq 0$, 
where $h(x)$ is continuously differentiable. The robot should not exceed this safe set throughout the entire task execution. More formally, with the robot dynamics in (\ref{eqn::control_affine_dynamics}), we want to ensure that with proper gains $u$ selection, $x$ remains in the safe set.

According to Nagumo’s theorem~\cite{blanchini1999set} and Zeroing Barrier Functions (Proposition 1 in~\cite{ames2016control}), a sufficient condition to ensure $h(x)\geq0$ is given as follows,
\begin{equation}
     h^\prime(x) = \dot{h}(x) + \gamma h(x) \geq 0 
\label{eqn_Nagumo_theorem}
\end{equation}
where $\dot{h}$ is the time derivative of $h(x)$ given in (\ref{eqn::lie_derivative}) and $\gamma > 0$ is an extended class $\kappa$ function suggested by Remark 6 in~\cite{ames2016control}.
\begin{equation}
    \dot{h}(x) = \frac{\partial h}{\partial x}^T \cdot \frac{dx}{dt} = \frac{\partial h}{\partial x}^T\cdot[f(x)+g(x)u]
\label{eqn::lie_derivative}
\end{equation}

Therefore, ensuring $h^\prime(x) \geq 0$ is sufficient to guarantee the robot stays in the safe set, and we can construct the constraint on $u$.
In this paper, we consider robot collision constraint $h(x) = h_p(e)$, which only depends on the robot position.
Since, the first element of $g(x)$ in the robot dynamics~(\ref{eqn::control_affine_dynamics}) is zero (relative degree greater than 1),
the impedance gains $u$ do not show up in $h^\prime(x)$ as shown in Appendix~A , we cannot directly obtain the constraints on $u$ to regulate the robot. Similarly, we note that $h^\prime(x) \geq 0$ is ensured if $\dot{h}^\prime(x) + \gamma h^\prime(x) \geq 0$. This condition results in~(\ref{eqn::collision_avoidance_CBF}) as the final collision constraint. The derivation is provided in the Appendix~A.
\begin{align}
    S(x) =& \bigg{\{} u \in \mathbb{R}_{18}^+ : \frac{\partial^2 h_p}{\partial e^2}\dot{e}^2 + \frac{\partial h_p}{\partial e}[g_2(x)u+2\gamma\dot{e}]
    \nonumber \\
    &+\gamma^2h_p(e)  \geq 0 \bigg{\}}
\label{eqn::collision_avoidance_CBF}
\end{align}

\subsection{Safe OnGO-VIC Algorithm}
The goal of the proposed Safe OnGO-VIC is to find optimal impedance gains and ensure safety. Since safety has the highest priority, we propose a framework that consists of two-level optimizations, a high-frequency safety optimization and a low-frequency gain optimization.

For the low-frequency gain optimization in (\ref{eqn_low_frequency}), the goal is to obtain the optimal impedance gains based on the online collected force data. The objective function is to minimize FITAVE in (\ref{eqn::ITAE_loss}) to achieve smooth behaviors.
\begin{equation}
\begin{aligned}
\min_{\bm{u} \in \mathbb{R}_{18}^+} \quad &  FITAVE
\label{eqn_low_frequency}
\end{aligned}
\end{equation}

For the high-frequency safety optimization, the goal is to keep the robot satisfying the safety constraints by changing impedance gains. Thus, we include the collision constraints that are shown in (\ref{eqn_high_frequency}). 
The objective function is designed to minimize the change of gains, where $u^{*}$ is the optimal gain values obtained from the previous low-frequency optimization step. 
\begin{equation}
\begin{aligned}
\min_{u \in \mathbb{R}_{18}^+} \quad &  \|u-u^*\|^2\\
\textrm{s.t.} \quad  & u \in S(x)\\
\label{eqn_high_frequency}
\end{aligned}
\end{equation}

The high frequency safety optimization (\ref{eqn_high_frequency}) is convex, while the low frequency gain optimization (\ref{eqn_low_frequency}) is not because of the environment force in $g_2(x)$. In practice, both problems can be tackled by a Sequential Quadratic Programming (SQP) solver~\cite{boggs1995sequential}.
After solving the optimization, impedance gains can be recovered by:
\begin{equation}
\begin{aligned}
    M &= diag\{u_{13},\cdots,u_{18}\}^{-1} \\
    K_d &= M \cdot diag\{u_1,\cdots,u_6\}  \\
    K_p &=  M \cdot diag\{u_7,\cdots,u_{12}\} \\
\end{aligned}
\label{eqn::gain_recover}
\end{equation}

The whole algorithm is illustrated in Alg. 1. At runtime, gains are initialized with random positive values. At each sampling interval, the high-frequency optimization is calculated based on the current robot states, and impedance gains are recovered by (\ref{eqn::gain_recover}). At the same time, the force sensor data is collected. Every $T$ second, the low-frequency gain adaptation optimization calculates the optimal impedance gain $u^*$ and sends it to the impedance controller.

\begin{algorithm}
\caption{Safe OnGO-VIC}
\begin{algorithmic}[1] 
\label{alg_Ongovic}
\REQUIRE Initialize $u$
\WHILE{task not terminated}
    \IF {every $T$ seconds}
        \STATE $u^* \leftarrow$ Low-frequency optimization in (\ref{eqn_low_frequency})
        \STATE $M, K_p, K_d \leftarrow$ Recover impedance gain by (\ref{eqn::gain_recover})
    \ENDIF
    \STATE $u \leftarrow$ High-frequency optimization in (\ref{eqn_high_frequency})
    \STATE $M, K_p, K_d \leftarrow$ Recover impedance gain by (\ref{eqn::gain_recover})
    \STATE $\{F\} \leftarrow$ Collect force sensor data
    \ENDWHILE

\end{algorithmic}
\end{algorithm}

\section{Experiments}
\label{EXPERIMENTS}

We test the proposed Safe OnGO-VIC in three tasks: collision avoidance, surface contact, and board wiping. 
The proposed method is benchmarked with two baselines, 1) conventional constant-gain impedance control~(CGIC) and 2) an advanced adaptive VIC (AVIC) proposed in~\cite{duan2018adaptive}.

\subsection{Experiment Setup}
\begin{figure}
    \centering
    \includegraphics[width=250pt]{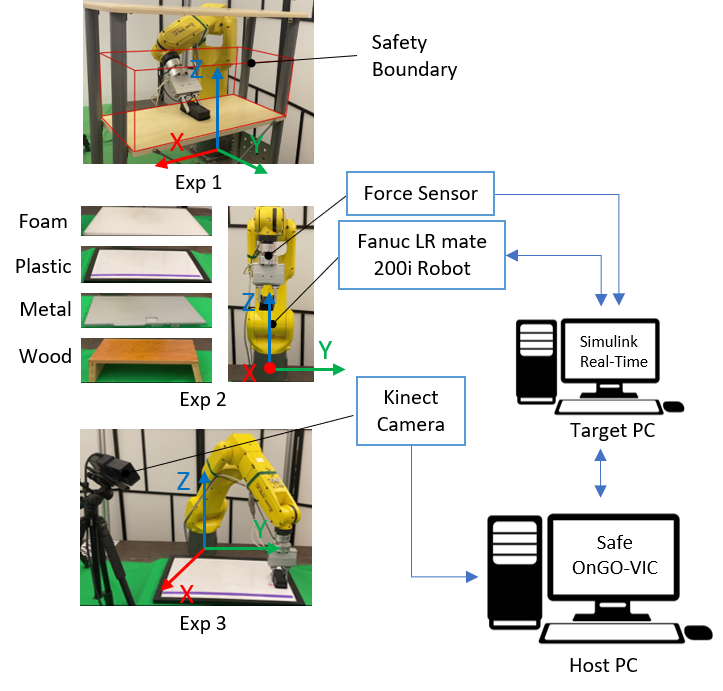}
    \caption{Experiment setup for three real-world experiments}
    \label{fig:setup}
\end{figure}

As shown in Fig.~\ref{fig:setup}, we use a 6-DOF FANUC LR Mate 200iD robot to validate the proposed method. The coordinate system of the robot is shown in Fig.~\ref{fig:setup}.
A Microsoft Kinect 2.0 is utilized to monitor the environment. The robot is equipped with an ATI Mini 45 F/T sensor to measure the external wrench. The Cartesian space impedance control law (\ref{impdeancecontrollaw}) is implemented in Simulink Real-Time on a target PC to control the robot. 
The proposed Safe OnGO-VIC is programmed on a host PC with a communication frequency of $125$Hz.

\subsection{Exp1: Safety Guarantee in Constrained Environments}
First, we evaluate the effectiveness of the high-frequency safety optimization in a constrained environment, where manipulation tasks should be accomplished within a safe region. 

As shown in Fig. \ref{fig:setup}, the robot picks an object on the shelf. While the human, as a random disturbance, disturbs the robot in different directions. We want to show that Safe OnGO-VIC is able to optimize impedance gains in order to stay safe.
The constraints in (\ref{eqn::safety_constraint}) is to enable the robot to stay within a `safety zone' and is enforced at $125Hz$,
\begin{equation}
\begin{aligned}
    h_1(x) & = d_{ub} - e(t) \geq 0 \\
    h_2(x) & = e(t) - d_{lb} \geq 0
    \label{eqn::safety_constraint}
\end{aligned}
\end{equation}
where $d_{lb}, d_{ub} \in \mathbb{R}^3$ are the safety boundaries.

\begin{figure}[http]
  \centering
   \includegraphics[width=223pt]{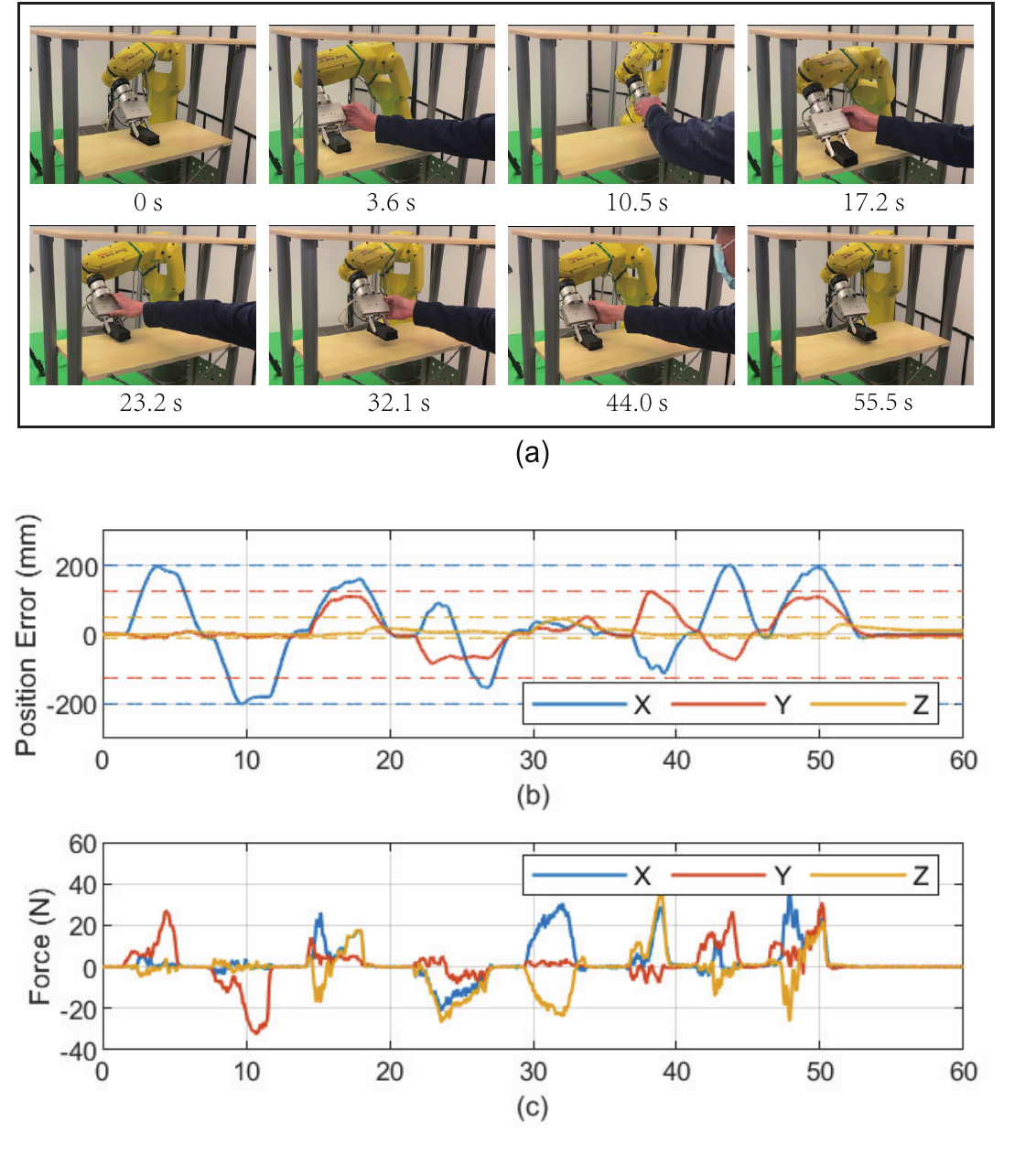}
   \caption{Exp1 with Safe OnGO-VIC: (a) shows snapshots of human disturbance. Solid lines in (b) stand for the robot trajectory, and the dashed lines are designed safety bound. The force applied by a human is recorded in (c).}
   \label{fig:exp1_proposed_method}
   \end{figure}

Fig. \ref{fig:exp1_proposed_method}~(a) shows the snapshots of the experiments. In the beginning, the robot is away from the safety boundary, and impedance gains $K_p$, $K_d$ and $M$ are chosen to achieve a smooth behavior under external force. After humans drag the end-effector towards the boundary, the robot is aware of the danger and quickly adapts the $K_p$, $K_d$, and $M$ to resist the disturbance and stay inside the safe region.

As a comparison, the results of the constant gain impedance control law are provided on our website. The results show that they are not able to stay within the safety region and avoid potential collisions with obstacles. 

\subsection{Exp2: Online Gain Adaptation in Unknown Environments}
\label{exp2}

\begin{table*}[hp]  
    \centering
    \caption{Experiment 2: Performance Comparison}  
    \label{tab:table1}
   \begin{tabular}{|c|c|c|c|c|c|c|c|c|c|}
  \cline{1-10} 
   \multicolumn{1}{|c|} {\multirow{2}{*}{}} & \multicolumn{3}{c|}{Plastic Board} &  \multicolumn{3}{c|}{Metal Board}  &  \multicolumn{3}{c|}{Wood Board} \\ 
    \cline{2-10}
    \multicolumn{1}{|c|}{}&CGIC & AVIC & Proposed &CGIC & AVIC & Proposed & CGIC & AVIC & Proposed\\
    \cline{1-10}         
   \multirow{1}{*}{Approaching Time (s)}  & $\bm{2.91}$ & $4.33$& $3.43$ & $\bm{2.92}$& $3.78$& $3.30$ & $\bm{2.29}$   & $4.18$    & $2.67$\\
    \cline{1-10} 
    \multirow{1}{*}{Settling Time of Position Error (s)}  & $> 4.47$& $16.54$& $\bm{1.85}$ & $\bm{2.01}$& $10.14$& $2.50$ & $> 6.59$   & $15.10$    & $\bm{1.39}$\\
   \cline{1-10} 
     \multirow{1}{*}{Variance of External Force at Steady State ($N^2$)}  & N/A& $6.52$& $\bm{0.12}$ & $3.54$& $2.21$& $\bm{0.14}$ & N/A   & $7.41$    & $\bm{0.09}$\\
    \cline{1-10} 
    \end{tabular}
    \end{table*}

\begin{figure*}[http]
  \centering
   \includegraphics[width=480pt]{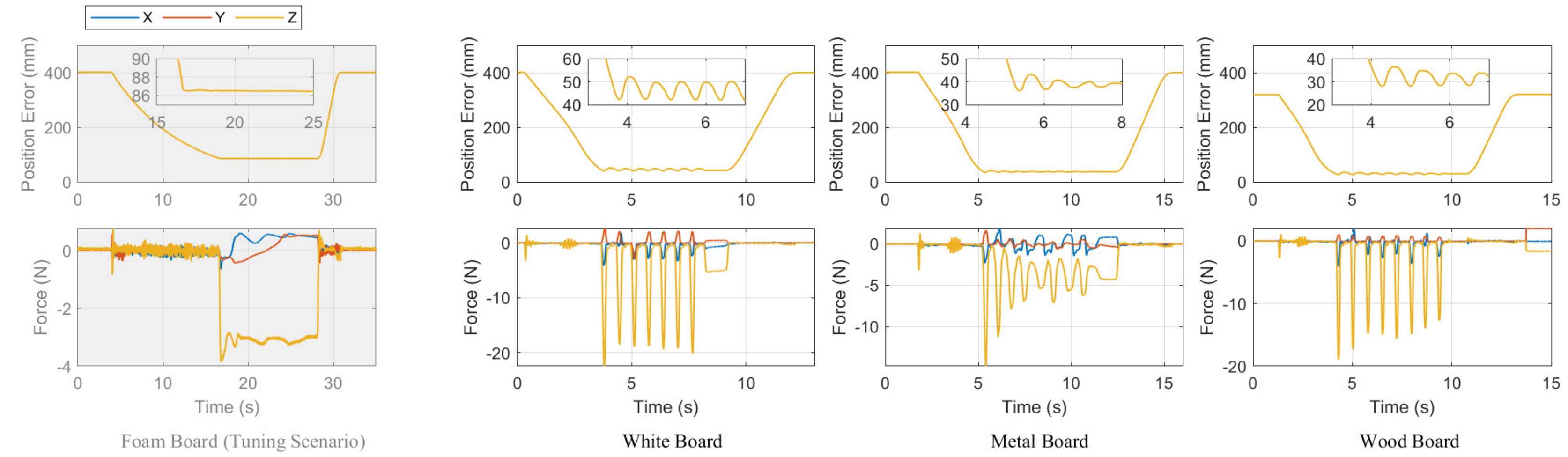}
   \caption{Results of Exp2 with CGIC}
   \label{fig:exp2_constant_gain_method}
   \end{figure*}

\begin{figure*}[http]
  \centering
   \includegraphics[width=480pt]{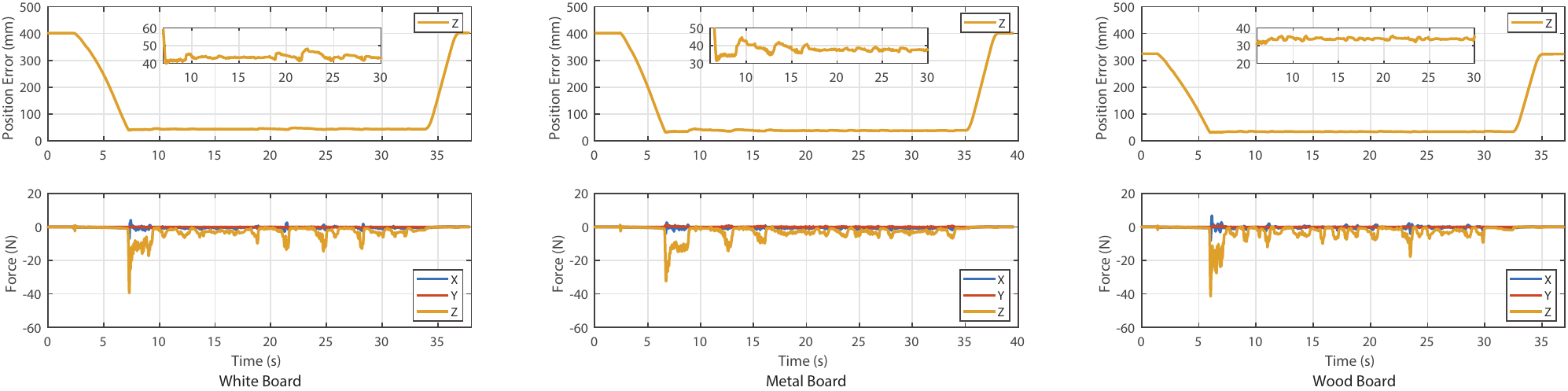}
   \caption{Results of Exp2 with AVIC}
   \label{fig:exp2_AVIC}
   \end{figure*}

\begin{figure*}[http]
  \centering
   \includegraphics[width=480pt]{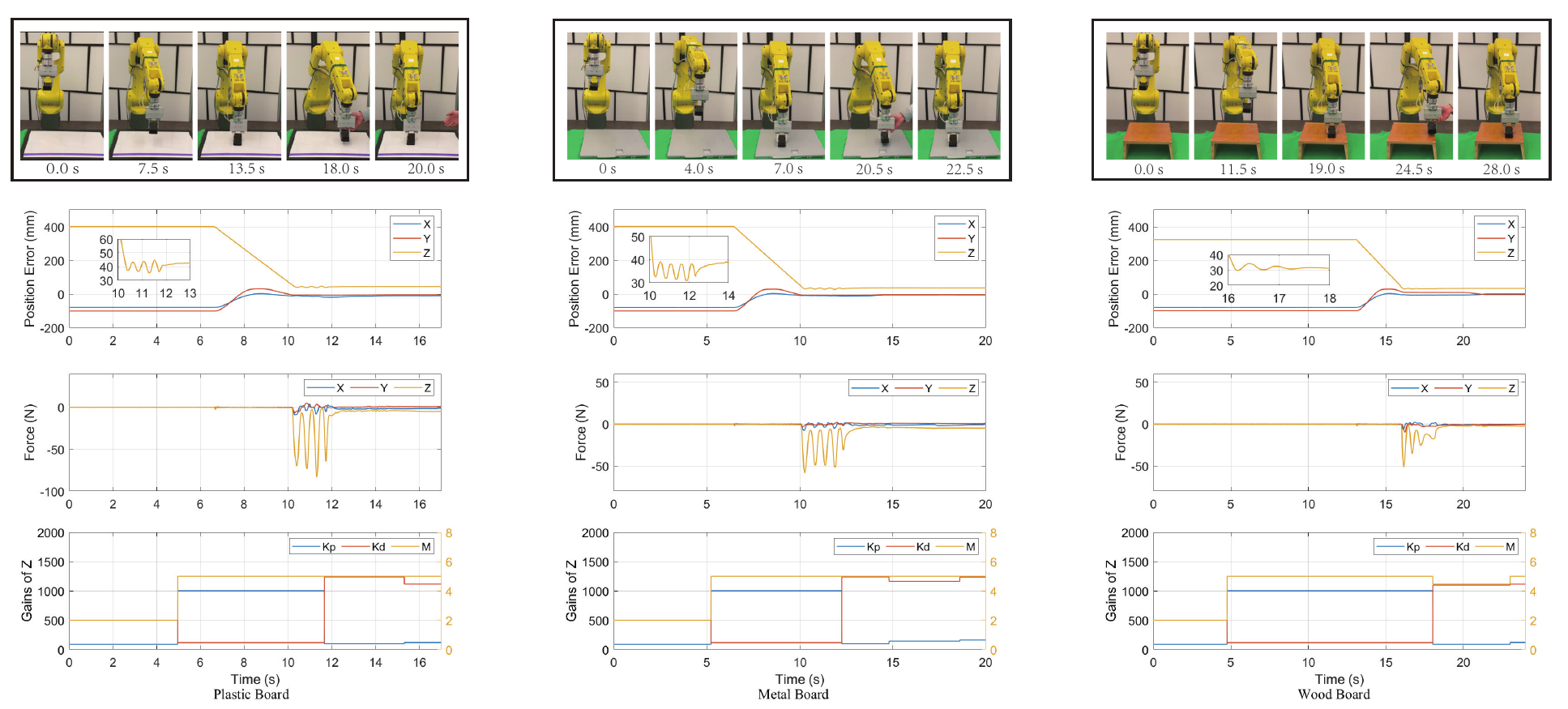}
   \caption{Results of Exp2 with Safe OnGO-VIC }
   \label{fig:exp2_proposed_method_WB}
   \end{figure*}

The experiment is to let the proposed algorithm obtain the optimal impedance gain (without human tuning) to make the robot end-effector smoothly contact with different surfaces, such as plastic, metal, and wood boards (as shown in Fig.~\ref{fig:setup}). 

We first test the performance of the constant gain baseline (CGIC).
As shown in Fig.~\ref{fig:exp2_constant_gain_method}, parameters of the constant gain method are tuned to the best performance on a foam surface and then transferred to three test surfaces. However, the constant gain baseline cannot achieve satisfactory results in other scenarios due to the lack of generalizability. Furthermore, the gain value does not reflect the change of the environment characteristics, and therefore, it results in the oscillation of the end-effector.

For the adaptive gain baseline (AVIC), the initial impedance gains are the same as CGIC. During the task execution, the control law adapts the damping term according to the current force value. From the result in Fig.~\ref{fig:exp2_AVIC}, the adaptive gain baseline performs better than the constant gain. However, it takes a long time to reach the steady-state and also results in some oscillation behaviors.

The proposed Safe OnGO-VIC algorithm does not require any offline tuning, and it is randomly initialized with some gain values. As shown in Fig.~\ref{fig:exp2_proposed_method_WB} and the videos, the robot initially contacts the board hardly and results in clear oscillations. While the robot measures the contact, the gain adaptation optimization is aware of the change of environment characteristics and updates the impedance values immediately to make contact smoother. From the figures and the videos, we can observe that after one step of update, the robot stops oscillating, and the force profile immediately becomes stable.
The low-frequency optimization collects the force sensor data for every $3$ second, and the optimization takes about $0.6$ seconds to solve. Moreover, after successfully adapting to the surface, the human lifts the robot to some random positions (at $18$~s, $23$~s, $28$~s, and $37$~s in the video on our website), the robot can recover to the desired position smoothly using the obtained optimal impedance gain. Similarly, the results on the metal board and wood board are illustrated in Fig.~\ref{fig:exp2_proposed_method_WB}~(b) and~(c), respectively.
According to the force curve, the results also reveal that our method has the smallest force ripple in the steady-state compared with CGIC and AVIC in all the scenarios. 

Table~\ref{tab:table1} summarizes the performance of each algorithm and quantitatively compares the results. 
It indicates that the proposed Safe OnGO-VIC algorithm can optimize the performance after the robot contacts the surfaces, achieving the shortest settling time and the smallest force variance at the steady state. Moreover, the proposed algorithm is able to approach the surface rapidly with shorter approaching times compared with AVIC. As for the constant gain method, the robot's position error cannot converge to the steady state on plastic and wood boards for a long time. These numerical results prove the advantage of the proposed method over CGIC and AVIC again.

 \begin{figure}[http]
  \centering
   \includegraphics[width=250pt]{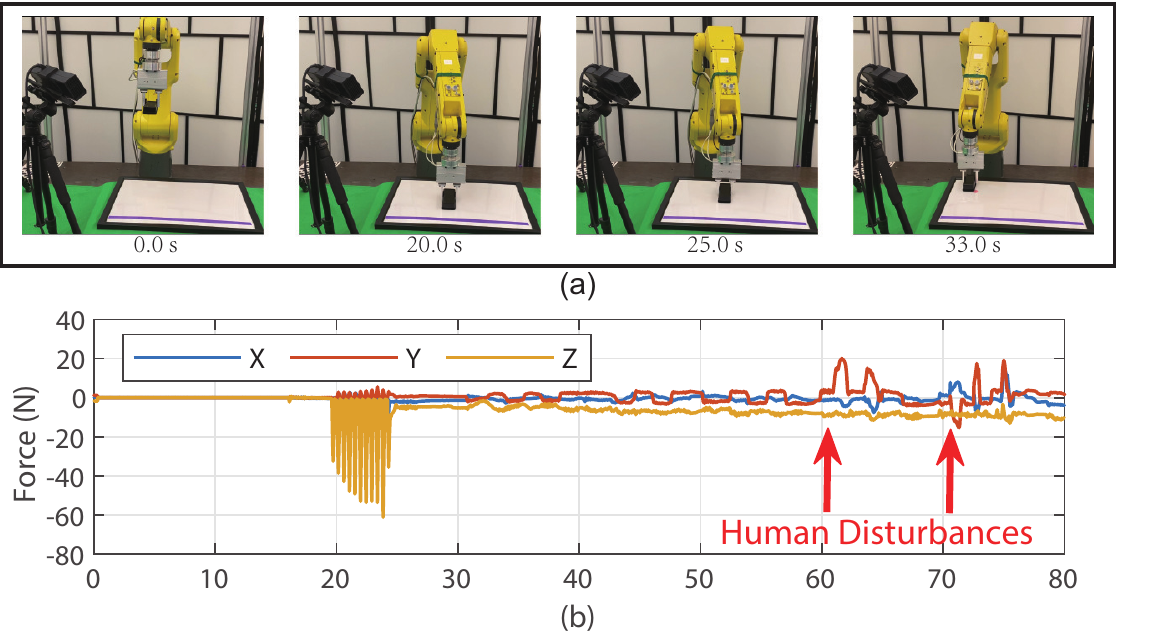}
   \caption{Exp3 with Safe OnGO-VIC}
   \label{fig:exp3}
   \end{figure}

\subsection{Exp3: Contact Rich Manipulation Task - Surface Wiping}
The experiment's goal is to provide an application of the proposed method (low-frequency and high-frequency) and verify its effectiveness and robustness further. As shown in Fig. \ref{fig:setup}, the robot contacts with the board and wipes the ``stain" without goes out to a predefined unsafe area (denotes by the purple line on the board). A Microsoft Kinect 2.0 is utilized to monitor the board and sends the ``stain" location to the controller as the reference position. 

Fig.~\ref{fig:exp3} illustrates the change of the robot position and impedance gains.
Similar to the results of Exp2, the robot finds optimal impedance gains after a few steps of adaptation, and when the human draws some ``markers'' on the board, the camera detects the position and sends it to the controller as a new reference point. At the same time, the gain adaptation optimization constantly collects the force data and optimizes impedance gains. The success of this task with the proposed method proves that our method can obtain the optimal impedance gain for different tasks online while guaranteeing safety at the same time.

\section{Conclusion}
\label{conclusion}
We proposed Safe OnGO-VIC, an efficient online gain optimization algorithm, for variable impedance control. The relation between impedance gains and robot behavior is explicitly constructed as a control-affine system. Based on that, an optimization problem is formulated to optimize impedance gains in real-time based on the online collected force data. Moreover, collision constraint is embedded into the framework to ensure safety. A series of experiments demonstrates the performance of the proposed method, and the comparison with the constant gain baseline and an adaptive gain baseline also indicates that the proposed method is more effective and robust to different unknown scenarios.

For future work, we plan to incorporate the regulation of contact force into the objective function to make the robot apply a constant force to the environment. Moreover, we will test the proposed algorithm in other contact-rich manipulation tasks, such as polishing and grinding.

\section{Appendix}
\label{appendix}
\subsection{Derivation of the Collision Constraint}
For collision constraint $h(x) = h_p(e)$ only on robot position, we have 
\begin{equation}
    \frac{\partial h}{\partial x} = \begin{bmatrix}
        \frac{\partial h_p}{\partial e}\\
        0
    \end{bmatrix}
\end{equation}
Substituting the above equation and the robot dynamics (\ref{eqn::control_affine_dynamics}) to (\ref{eqn_Nagumo_theorem}), we can obtain
\begin{equation}
    \begin{aligned}
        h^{\prime}(x) & = \dot{h}(x) + \gamma h(x) \\
        & = \frac{\partial h}{\partial x}^T\cdot[f(x)+g(x)u] + \gamma h(x)\\
        & = \begin{bmatrix}
        \frac{\partial h_p}{\partial e}\\
        0
    \end{bmatrix}^T \cdot
    \begin{bmatrix}
        \dot{e}\\
        g_2(x)u
    \end{bmatrix} + \gamma h_p(e)\\
    & = \frac{\partial h_p}{\partial e} \dot{e} + \gamma h_p(e)
    \end{aligned}
\end{equation}

Therefore, ensuring $h^\prime(x) \geq 0$ is sufficient to guarantee the robot stays in the safe set. However, the impedance gains $u$ do not show up in the above equation (relative degree greater than 1), we cannot directly obtain the constraints on $u$ to regulate the robot. Therefore, we apply the condition (\ref{eqn_Nagumo_theorem}) once more on $h^\prime(x)$ to get the collision constraint. 
\begin{equation}
    \begin{aligned}
        \dot{h^\prime}(x) + \gamma h^\prime(x) & = \frac{\partial h^\prime}{\partial x}^T\cdot\frac{dx}{dt} + \gamma h^\prime(x) \\
        & = \begin{bmatrix}
            \frac{\partial}{\partial e}[\frac{\partial h_p}{\partial e} \dot{e} + \gamma h_p(e)]\\
            \frac{\partial}{\partial \dot{e}} [\frac{\partial h_p}{\partial e} \dot{e} + \gamma h_p(e)]
        \end{bmatrix}^T\cdot\frac{dx}{dt} + \gamma h^\prime(x)\\
        & = \begin{bmatrix}
            \frac{\partial^2 h_p}{\partial e^2}\dot{e} + \gamma \frac{\partial h_p}{\partial e} \\
            \frac{\partial h_p}{\partial e}
        \end{bmatrix}^T\cdot\frac{dx}{dt} + \gamma h^\prime(x)\\
        & = \frac{\partial^2 h_p}{\partial e^2}\dot{e}^2 + \frac{\partial h_p}{\partial e}[g_2(x)u+2\gamma\dot{e}]+\gamma^2h_p(e)
    \end{aligned}\nonumber 
\end{equation}

\bibliographystyle{IEEEtran}
\bibliography{Reference}	

\begin{thebibliography}{10}
\providecommand{\url}[1]{#1}
\csname url@samestyle\endcsname
\providecommand{\newblock}{\relax}
\providecommand{\bibinfo}[2]{#2}
\providecommand{\BIBentrySTDinterwordspacing}{\spaceskip=0pt\relax}
\providecommand{\BIBentryALTinterwordstretchfactor}{4}
\providecommand{\BIBentryALTinterwordspacing}{\spaceskip=\fontdimen2\font plus
\BIBentryALTinterwordstretchfactor\fontdimen3\font minus
  \fontdimen4\font\relax}
\providecommand{\BIBforeignlanguage}[2]{{%
\expandafter\ifx\csname l@#1\endcsname\relax
\typeout{** WARNING: IEEEtran.bst: No hyphenation pattern has been}%
\typeout{** loaded for the language `#1'. Using the pattern for}%
\typeout{** the default language instead.}%
\else
\language=\csname l@#1\endcsname
\fi
#2}}
\providecommand{\BIBdecl}{\relax}
\BIBdecl

\bibitem{muratore2019self}
L.~Muratore, A.~Laurenzi, and N.~G. Tsagarakis, ``A self-modulated impedance
  multimodal interaction framework for human-robot collaboration,'' in
  \emph{2019 International Conference on Robotics and Automation (ICRA)}.\hskip
  1em plus 0.5em minus 0.4em\relax IEEE, 2019, pp. 4998--5004.

\bibitem{johannsmeier2019framework}
L.~Johannsmeier, M.~Gerchow, and S.~Haddadin, ``A framework for robot
  manipulation: Skill formalism, meta learning and adaptive control,'' in
  \emph{2019 International Conference on Robotics and Automation (ICRA)}.\hskip
  1em plus 0.5em minus 0.4em\relax IEEE, 2019, pp. 5844--5850.

\bibitem{peternel2015human}
L.~Peternel, T.~Petri{\v{c}}, and J.~Babi{\v{c}}, ``Human-in-the-loop approach
  for teaching robot assembly tasks using impedance control interface,'' in
  \emph{2015 IEEE international conference on robotics and automation
  (ICRA)}.\hskip 1em plus 0.5em minus 0.4em\relax IEEE, 2015, pp. 1497--1502.

\bibitem{abu2018force}
F.~J. Abu-Dakka, L.~Rozo, and D.~G. Caldwell, ``Force-based learning of
  variable impedance skills for robotic manipulation,'' in \emph{2018 IEEE-RAS
  18th International Conference on Humanoid Robots (Humanoids)}.\hskip 1em plus
  0.5em minus 0.4em\relax IEEE, 2018, pp. 1--9.

\bibitem{zhang2021learning}
X.~Zhang, L.~Sun, Z.~Kuang, and M.~Tomizuka, ``Learning variable impedance
  control via inverse reinforcement learning for force-related tasks,''
  \emph{arXiv preprint arXiv:2102.06838}, 2021.

\bibitem{li2017adaptive}
Z.~Li, J.~Liu, Z.~Huang, Y.~Peng, H.~Pu, and L.~Ding, ``Adaptive impedance
  control of human--robot cooperation using reinforcement learning,''
  \emph{IEEE Transactions on Industrial Electronics}, vol.~64, no.~10, pp.
  8013--8022, 2017.

\bibitem{martin2019variable}
R.~Mart{\'\i}n-Mart{\'\i}n, M.~A. Lee, R.~Gardner, S.~Savarese, J.~Bohg, and
  A.~Garg, ``Variable impedance control in end-effector space: An action space
  for reinforcement learning in contact-rich tasks,'' \emph{arXiv preprint
  arXiv:1906.08880}, 2019.

\bibitem{rey2018learning}
J.~Rey, K.~Kronander, F.~Farshidian, J.~Buchli, and A.~Billard, ``Learning
  motions from demonstrations and rewards with time-invariant dynamical systems
  based policies,'' \emph{Autonomous Robots}, vol.~42, no.~1, pp. 45--64, 2018.

\bibitem{mehdi2011impedance}
H.~Mehdi and O.~Boubaker, ``Impedance controller tuned by particle swarm
  optimization for robotic arms,'' \emph{International Journal of Advanced
  Robotic Systems}, vol.~8, no.~5, p.~57, 2011.

\bibitem{lahr2017adjustable}
G.~J. Lahr, H.~B. Garcia, J.~O. Savazzi, C.~B. Moretti, R.~V. Aroca, L.~M.
  Pedro, G.~F. Barbosa, and G.~A. Caurin, ``Adjustable interaction control
  using genetic algorithm for enhanced coupled dynamics in tool-part contact,''
  in \emph{2017 IEEE/RSJ International Conference on Intelligent Robots and
  Systems (IROS)}.\hskip 1em plus 0.5em minus 0.4em\relax IEEE, 2017, pp.
  1630--1635.

\bibitem{song2017impedance}
P.~Song, Y.~Yu, and X.~Zhang, ``Impedance control of robots: an overview,'' in
  \emph{2017 2nd international conference on cybernetics, robotics and control
  (CRC)}.\hskip 1em plus 0.5em minus 0.4em\relax IEEE, 2017, pp. 51--55.

\bibitem{kronander2016stability}
K.~Kronander and A.~Billard, ``Stability considerations for variable impedance
  control,'' \emph{IEEE Transactions on Robotics}, vol.~32, no.~5, pp.
  1298--1305, 2016.

\bibitem{lawrence1988impedance}
D.~A. Lawrence, ``Impedance control stability properties in common
  implementations,'' in \emph{Proceedings. 1988 IEEE International Conference
  on Robotics and Automation}.\hskip 1em plus 0.5em minus 0.4em\relax IEEE,
  1988, pp. 1185--1190.

\bibitem{seraji1997force}
H.~Seraji and R.~Colbaugh, ``Force tracking in impedance control,'' \emph{The
  International Journal of Robotics Research}, vol.~16, no.~1, pp. 97--117,
  1997.

\bibitem{martins2005tuning}
F.~G. Martins, ``Tuning pid controllers using the itae criterion,''
  \emph{International Journal of Engineering Education}, vol.~21, no.~5, p.
  867, 2005.

\bibitem{blanchini1999set}
F.~Blanchini, ``Set invariance in control,'' \emph{Automatica}, vol.~35,
  no.~11, pp. 1747--1767, 1999.

\bibitem{ames2016control}
A.~D. Ames, X.~Xu, J.~W. Grizzle, and P.~Tabuada, ``Control barrier function
  based quadratic programs for safety critical systems,'' \emph{IEEE
  Transactions on Automatic Control}, vol.~62, no.~8, pp. 3861--3876, 2016.

\bibitem{boggs1995sequential}
P.~T. Boggs and J.~W. Tolle, ``Sequential quadratic programming,'' \emph{Acta
  numerica}, vol.~4, pp. 1--51, 1995.

\bibitem{duan2018adaptive}
J.~Duan, Y.~Gan, M.~Chen, and X.~Dai, ``Adaptive variable impedance control for
  dynamic contact force tracking in uncertain environment,'' \emph{Robotics and
  Autonomous Systems}, vol. 102, pp. 54--65, 2018.

\end{thebibliography}
\vspace{12pt}
\end{document}